# VRFP: On-the-fly Video Retrieval using Web Images and Fast Fisher Vector Products

Xintong Han*, Bharat Singh*, Vlad I. Morariu, Larry S. Davis, *Fellow, IEEE*

***Abstract*—VRFP is a real-time video retrieval framework based on short text input queries, which obtains weakly labeled training images from the web after the query is known. The retrieved web images representing the query and each database video are treated as unordered collections of images, and each collection is represented using a single Fisher Vector built on CNN features. Our experiments show that a Fisher Vector is robust to noise present in web images and compares favorably in terms of accuracy to other standard representations. While a Fisher Vector can be constructed efficiently for a new query, matching against the test set is slow due to its high dimensionality. To perform matching in real-time, we present a lossless algorithm that accelerates the inner product computation between high dimensional Fisher Vectors. We prove that the expected number of multiplications required decreases quadratically with the sparsity of Fisher Vectors. We are not only able to construct and apply query models in real-time, but with the help of a simple re-ranking scheme, we also outperform state-of-the-art automatic retrieval methods by a significant margin on TRECVID MED13 (3.5%), MED14 (1.3%) and CCV datasets (5.2%). We also provide a direct comparison on standard datasets between two different paradigms for automatic video retrieval—zero-shot learning and on-the-fly retrieval.**

***Index Terms*—Video Retrieval; Web-Based Retrieval; Fisher Vectors; Fast inner products**

## I. Introduction

With the shrinking cost of cloud storage provided by services such as iCloud, Dropbox, and Google Drive, etc., users have started collecting and storing personal multimedia data on a large scale. A typical user's media library may contain thousands of videos after a few years. However, these videos are usually not labeled and it eventually becomes very tedious to search through them for specific content. Therefore, our goal is to develop a search engine that performs visual search in a user's personal media library efficiently. A user inputs a text query, and our system ranks the videos according to their similarity to the query, learning the mapping between linguistic and visual representations on-the-fly. Like any search engine, the input is only a short text query, rather than a detailed description.

This work is supported by the Intelligence Advanced Research Projects Activity (IARPA) via the Department of Interior National Business Center contract number D11PC20071. The U.S. Government is authorized to reproduce and distribute reprints for Governmental purposes not with standing any copyright annotation thereon. The views and conclusions contained herein are those of the authors and should not be interpreted as necessarily representing the official policies or endorsements, either expressed or implied, of IARPA, DoI/NBC, or the U.S. Government.

*The first two authors contributed equally to this paper.

X. Han, B. Singh, V. I. Morariu, and L. S. Davis are with Center for Automation Research, University of Maryland, College Park, MD, 20742. E-mail: {xintong, bharat, morariu, lsd}@umiacs.umd.edu

The number of potential text queries is so large that it is impractical to enumerate all potential queries and train a visual model for each query. Instead, two techniques are commonly used to efficiently handle arbitrary queries: zero-shot learning and on-the-fly training. Zero-shot methods obtain a visual model for a query without requiring training samples for that query [7], [9], [10], [19], [26], [34], [50], [51], [54]. On-the-fly methods build a model using training samples that are collected only after the query is known [2], [4]–[6], [15].

Zero-shot frameworks are well-suited for tasks like video retrieval because of their computational efficiency. They generally apply a pre-defined set of concept detectors (or a concept bank) to a database of videos a-priori, and use a known mapping function (like distance in word2vec [36] space) to match a query to the set of concept detectors in semantic space. After selecting the nearest concepts in semantic space, a weighted average of concept detectors (based on the semantic similarity to the query) is used to obtain a ranked list of videos. Since the concept detectors are applied on the database of videos a-priori and computing semantic similarity is fast, retrieval can be performed in real-time. However, zero-shot methods struggle to generalize to queries which are semantically *far* from the concept bank.

On-the-fly retrieval methods obtain their training data after the query is given (e.g., from the web), rather than relying on pre-trained concept banks. These methods more easily support arbitrary queries, as the amount of webly-supervised data available is orders of magnitude greater than any concept bank that could be reasonably constructed today. Further, with modern computing architectures like GPUs, these methods can achieve real-time performance [5]. On-the-fly methods have not previously been compared to zero-shot learning based algorithms for text-based video retrieval, so it is unclear if they perform better or worse in a similar setting. We present the results of an experimental study comparing such methods.

The representations of the database videos and search query, whether obtained by zero-shot or on-the-fly techniques, have a large impact on search efficiency and accuracy. In prior work [5]–[7], [45], web images were treated as independent images belonging to a class (or concept), on which classifiers like SVM are trained. Classifiers are applied to individual frames of a video (instead of a single feature vector representing the whole video), followed by some form of pooling (average/max). Applying classifiers to every frame is computationally costly [45]. An approximation can be made by average/max pooling the features of the video and applying the classifiers to the pooled features to reduce computational costs. However, this creates a mismatch between the feature representation on which the classifiers are trained and the feature representation to which they are applied, since training is performed on features of individual frames while testing is performed on features pooled over multiple frames.

We propose an on-the-fly retrieval approach that employs text-based web-image search to build a visual model and uses a Fisher Vector representation to compare the query to a video dataset efficiently without introducing a mismatch between training and testing features. Our work involves the following novel contributions:

- We propose to construct a *single* Fisher Vector representation for both web images and video frames. We compare this representation with a number of other query and video representations.
- We demonstrate that a simple inner product between Fisher Vectors constructed on CNN features of web images obtained from the query and database video frames is robust to noise present in web images and is an effective similarity measure between web images and videos.
- We propose an efficient matching algorithm that leverages the sparsity of high-dimensional Fisher Vectors induced by the significant correlation between web images and video frames. We prove that on expectation, the number of arithmetic operations decreases quadratically in terms of sparsity in Fisher Vectors.

The remainder of this paper is organized as follows. Section II discusses related work. Section III presents our on-the-fly retrieval approach. Section IV describes the Fast Fisher Vector Products algorithm for speeding up the process of video retrieval. We present the results of a comprehensive set of experiments on three datasets in Section V. Finally, Section VI concludes this paper.

## II. Related Work

Concept-based video representations have been widely used in the multimedia community to model complex queries for videos in a supervised setting [31], [32], [35], [55], [58]. Merler et al. [35] learn a bank of generic concept detectors from labeled web images to build an intermediate level semantic video representation. Queries are represented directly in terms of concepts, which are complementary to low-level visual descriptors. Mazloom et al. [32] choose the best set of concepts for a query from more than a thousand pre-trained concepts based on a feature selection algorithm. Zhang et al. [58] automatically construct a semantic-visual knowledge base using WordNet [37] and ImageNet [11]. Then, they select related semantic concepts for a query based on this knowledge base as an intermediate level representation.

Most video retrieval methods that do not require knowledge of the query at training time, and thus do not train for specific queries, also rely on concept-based video representations. These methods first pre-train a large set of concept detectors. The corresponding detector confidences on dataset videos are then used to form semantic video representations [9], [18], [33], [50]. Wu et al. [50] use off-the-shelf concept detectors and multimedia features to represent a video. Text and video features are then projected to a high-dimensional semantic space. Finally, video similarity scores are computed in this space to rank videos. By harvesting web videos and their descriptions, Habibian et al. [18] learn an embedding by jointly optimizing the semantic descriptiveness and the visual predictability of the embedding. Mazloom et al. [33] generate concept prototypes from a large web video collection, where each concept prototype contains a set of frames that are relevant to a semantic concept. Similarity between the text description and a video is measured by matching the video frames with the concept prototype dictionary.

However, it is difficult to decide which concepts should be trained without prior knowledge of the query. If the semantic gap between a query and the concept bank is large, methods that depend on a pre-defined concept bank will perform poorly. To reduce the semantic gap between the video description and the concept bank, some recent methods discover the concepts after the query is provided [7], [45]. Chen et al. [7] search the verb-noun pairs in the text query on Flickr, and select visually meaningful concepts based on the tags associated with Flickr images. Then, 2,000 detectors are trained using web images and applied to database videos. Based on [7], Singh et al. [45] build pair-concepts and select the relevant concepts by a series of concept pruning schemes.

Obtaining concepts after the query is provided ensures that they are semantically related, but, the associated concept detectors still need to be applied to almost every frame in the video database, which is computationally expensive [7], [45]. In contrast, on-the-fly retrieval methods do not have this problem, as they only collect web images for the search query [2], [4], [6], [15]. On-the-fly methods have been used for large scale object retrieval [4], [6], [15], face retrieval in videos [4], and place and logo recognition [2]. As time has progressed, search engines have become more advanced. As our experiments show, they can be effectively used for collecting training samples for complex queries, especially, when no visual prior is available. However, the method used to measure similarity between web images and videos impacts retrieval performance significantly. For on-the-fly retrieval, a linear SVM is typically trained [2], [4], [6], [15], as it is fast to train and predict. Unlike approaches that train SVMs to obtain a representation, we build a single Fisher Vector [39], [40] on CNN features of web images. We show that measuring similarity using inner products between Fisher Vectors of web images and dataset videos performs significantly better than training a linear SVM on web images or computing inner product between average pooled CNN features.

Fisher Vectors are a good representation for video frames, but they are high dimensional, so computing their inner products is significantly more expensive (around 20-50 times) than applying a linear SVM on average pooled CNN features of video frames. Methods have been previously developed for accelerating retrieval using Fisher Vectors or other high dimensional features [16], [21], [22], [43]. One of the most powerful techniques is product quantization (PQ) [20]. Here, each feature vector is decomposed into equal length sub-vectors, and a lookup table is constructed at query time to compute inner-products between two sub-vectors efficiently.

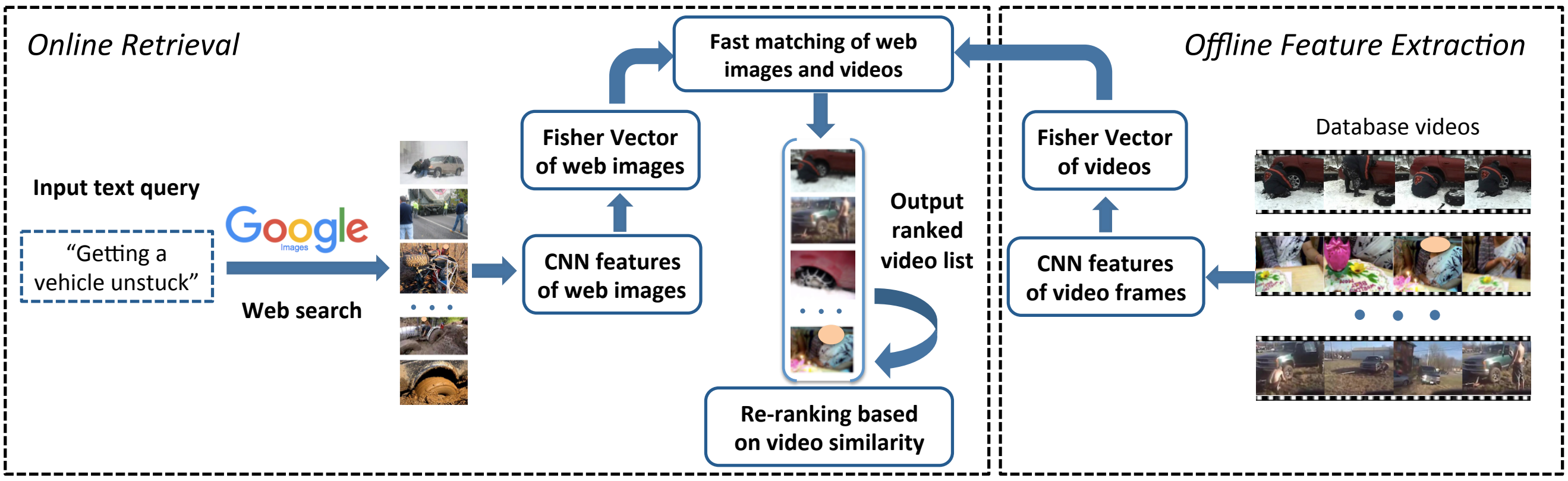

Fig. 1. VRFP summary. The approach contains two parts: offline feature extraction (right) and online video retrieval (left). During the offline step, Fisher Vectors of frame-level CNN features are built for database videos. Given a new text query (*Getting a vehicle unstuck*) as the input, an on-line visual representation is constructed efficiently via Fisher Vector encoding. Using an efficient matching function, we compute the similarity between the query and database videos to obtain a rank list. We use pseudo-relevance feedback to reduce the effects of domain shift between the web image and video domains, producing a re-ranked video list.

To further accelerate PQ, a High Variance Subspace First approach was proposed [56]. In this method, instead of computing distances to all the codewords, only the distance from high variance codewords is computed to reduce computation. However, these methods take a hit in performance as they are lossy compression techniques. Sparse matrix methods speed up inner product computation without compromising accuracy [57]. They avoid multiplying elements which are zero in one matrix, but it is not possible to take advantage of sparsity in both matrices simultaneously [57]. We present an algorithm that takes advantage of sparsity in both components of the inner product, avoiding any multiplication with a value of zero. We also propose a slight modification to this algorithm, to improve speedup, in a lossy setting.

Several methods speed up image and video retrieval using bag-of-words model built on low-level features like SIFT [28], [38], [46]. A Term Frequency-Inverse Document Frequency scheme performs fast matching between visual words of the query image/video and those of database images/videos [46]. A vocabulary tree based quantization scheme, which also performs indexing simultaneously, was shown to improve retrieval results [38]. In [28], a random forest is adopted to reduce the cost of bag-of-words quantization, and an SVM with fast Histogram Intersection kernel is used for retrieval. However, a Fisher Vector lies in a continuous high dimensional vector space. Therefore, approaches applicable to a bag-of-words model cannot be used to accelerate inner-products between Fisher-Vectors. To address this issue, a cascade architecture was proposed to accelerate inner-products between two vectors [6]. Since the first classifier needs to be applied on the complete dataset, it only achieves a 2x improvement in runtime.

Web images have been used for various tasks such as concept discovery [8], [12], event recognition [13], temporal localization of actions [47] etc. Many methods rely on relevance feedback [24] to mitigate the domain shift between web data and test set. Jiang et al. [24] use easy samples to perform re-ranking of the initial video list. To adapt detector weights, Tang et al. [48] gradually update the weights of web-trained detectors as they are applied to videos. Singh et al. [45] use the top ranked videos as positives, train a detector, and use it to re-rank dataset videos. In addition to addressing the domain shift in web data, many reranking approaches have also been utilized to boost the performance of image retrieval [2], [42], [44], [59], [60]. Shen et al. [44] and Arandjelovic et al. [2] re-rank the initial results by their spatial consistency. Since spatial consistency estimation is computationally costly, especially for videos, it is not suitable for our task. Zhang et al. [59] use a late fusion technique to update the initial rank list. Although some prior methods leverage K-nearest neighbors of the query image to perform reranking [2], [41], [42], [44] (see [60] for a comprehensive review), they can only be applied to visual word based image retrieval. We also investigate relevance feedback for re-ranking, which can be efficiently implemented and, as our experiments will show, leads to significant improvements.

## III. ON-THE-FLY RETRIEVAL APPROACH

VRFP is summarized in Fig. 1. The input to VRFP is a text query (*Getting a vehicle unstuck*). The output is a list of ranked videos where higher ranked videos are more relevant to the query (frames in the blue parentheses). As a pre-processing step, a Fisher Vector is constructed for every video in the dataset, as shown on the right of Fig. 1. For any given text query, web search is used to collect images relevant to the query that are then represented by a Fisher Vector of their CNN features. The similarity of each database video to the query is obtained by computing the similarity between the Fisher Vector representations of the query and each database video. As the visual representation is built from web images, there is a slight domain shift between the features of the database videos and web images. Thus, we construct a new visual representation for the query only on the basis of top ranked video features to re-rank the videos.

### *A. Compact Representations for Web Images and Videos*

The process of building a visual representation for the query needs to be efficient, as our goal is real-time video retrieval. The final representation should not depend on the number of images returned by web search or the number of concepts related to the query. Web search returns a set of images, and videos can be effectively represented as a set of frames, so we generate a compact representation for both videos and web images by treating them both as unordered image sets. In this section, we discuss methods for building compact representations of unordered image sets.

**Average Pooling**. A common approach to representing an image collection is to perform average pooling, i.e., compute an average feature vector from the set of images. Formally, the feature vector of a video or an image collection $X$ is $f_{avg}(X) = \frac{1}{N}\sum_{i=1}^{N} x_i$, where $N$ is the size of $X$, and $x_i$ is the feature vector of the $i^{th}$ frame or image. Finally, this feature vector is normalized by its $L_2$ norm.

**Max Pooling**. An alternative to average pooling, max pooling builds a representation by taking a dimension-wise maximum for frames in a video or an image collection, i.e., $f_{max}(X)^k = max_{i\in\{1,2,...,N\}} x_i^k$, where $N$ is the size of $X$, and $x_i^k$ is the $k^{th}$ dimension feature of the $i^{th}$ frame or image. Again, we apply $L_2$ normalization to the max pooled feature.

**Linear SVM**. A linear SVM [14] can be trained and used as a discriminative representation for web images. For a given query, a linear SVM is trained using web images collected for the query as positive samples and randomly sampled negatives. Then this SVM is applied to video frames for ranking the database videos. Linear SVMs have been widely used in several on-the-fly retrieval techniques [4], [6] for image retrieval as they offer a good compromise between speed and accuracy. However, as we are retrieving videos, applying the detector on every video frame would be computationally too expensive. Instead, we apply the trained SVM to an average pooled video representation.

**Fisher Vector (FV) Encoding**. Fisher Vector encoding first builds a $K$-component GMM model $(\mu_i, \sigma_i, w_i : i = 1, 2, ..., K)$ from training data, where $\mu_i, \sigma_i, w_i$ are the mean, diagonal covariance, and mixture weights for the $i^{th}$ component, respectively. Given a bag of features $\{x_1, x_2, ..., x_T\}$, its Fisher Vector is computed as:

$$\mathcal{G}_{\mu_i} = \frac{1}{T\sqrt{w_i}} \sum_{t=1}^{T} \gamma_t(i) \left( \frac{x_t - \mu_i}{\sigma_i} \right) \quad (1)$$

$$\mathcal{G}_{\sigma_i} = \frac{1}{T\sqrt{2w_i}} \sum_{t=1}^{T} \gamma_t(i) \left( \frac{(x_t - \mu_i)^2}{\sigma_i^2} - 1 \right) \quad (2)$$

where, $\gamma_t(i)$ is the posterior probability. Then all the $\mathcal{G}_{\mu_i}$ and $\mathcal{G}_{\sigma_i}$ are stacked to form the Fisher Vector. Following previous work [40], we compute a signed square-root on the Fisher Vector and then $L_2$ normalization.

**VLAD Encoding**. Similar to Fisher Vectors, VLAD encoding also aggregates a bag of features into a single high dimensional generative representation and has been shown to be effective for supervised video event retrieval [53]. VLAD first performs $K$-means clustering on the training data to obtain $K$ clustering centers $(\mu_i : i = 1, 2, ..., K)$. For the given features $\{x_1, x_2, ..., x_T\}$, it computes difference with the cluster centers and these features are stacked to construct a long vector:

$$\mathcal{G}_{\mu_i} = \sum_{t:NN(x_t)=\mu_i} (x_t - \mu_i) \quad (3)$$

where $NN(x_t) = \mu_i$ means $\mu_i$ is the nearest neighbor of $x_t$ among all cluster centers. Following [53], we also extend VLAD to VLAD-$k$, where $k$-nearest neighbors are used to encode the descriptor - $k$ is set to 3 in our experiments. Finally, signed square-rooting, intra-normalization, and $L_2$ normalization are applied on VLAD as in [3].

These pooling methods have been widely used to pool features of unordered sets of patches or images into a fixed length feature, e.g., pooling frame-level features for one video, pooling SIFT features in an image, or pooling motion features such as MBH in a video. It was recently shown that a VLAD representation [53] on CNN features obtains state-of-the-art results for event classification. We propose to build a single Fisher Vector for *web images* using their CNN representations. Unlike traditional linear SVM representations [4], [6] , which classify single images, a Fisher Vector captures the characteristics of the whole image collection.

### *B. Matching Web Images and Videos*

The above encoding schemes represent each query and video as a single feature vector. We perform video retrieval by computing the similarity between the query vector and each video vector. A simple way to measure similarity between the images and a video is to compute a cosine distance between their features; assuming features are $L_2$ normalized:

$$S_i = cosine(f_I, f_{V_i}) \quad (4)$$

where $f_I$, $f_{V_i}$ are the features of the image collection $I$ returned by web search and the frames comprising the video $V_i$, respectively. The database videos are ranked based on their similarity $S_i$ with $I$.

Although we use the Fisher Vector of CNN features as our representation, our matching process differs from previous work. In [53], a linear SVM is trained on Fisher Vectors/VLAD of training videos. However, during on-the-fly retrieval, we construct only one Fisher Vector from web images, so we do not have enough training samples to train an SVM. Therefore, we employ a nearest-neighbor approach for retrieval and show that this process can be accelerated without any loss in accuracy. This acceleration may not be possible if we use a linear SVM.

### *C. Re-ranking by Video Similarity*

Since the visual representation for the query is constructed from web images, there is a domain shift between video and image representations. Pseudo-relevance feedback is commonly used to deal with this type of domain shift. Pseudo relevance feedback involves, training an SVM on top ranked videos as positive samples and bottom ranked videos as negative samples [45]. Although other approaches have been proposed [24], [25], they take a few minutes to run at test time. Training an SVM may also take significant time, especially if it is trained on high dimensional features like Fisher Vectors. Moreover, most negative samples are not very informative due to the large diversity of background videos in the database.

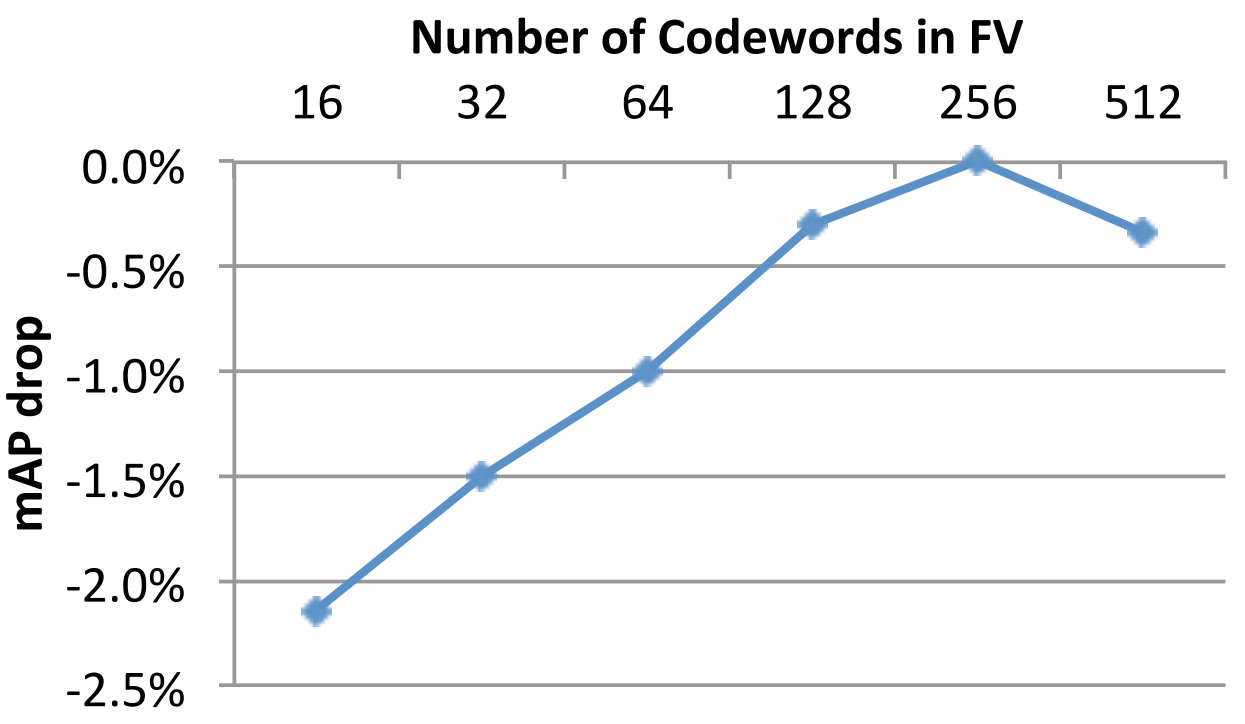

Fig. 2. Drop in performance (mAP) for the TRECVID MED13 data for different codeword sizes.

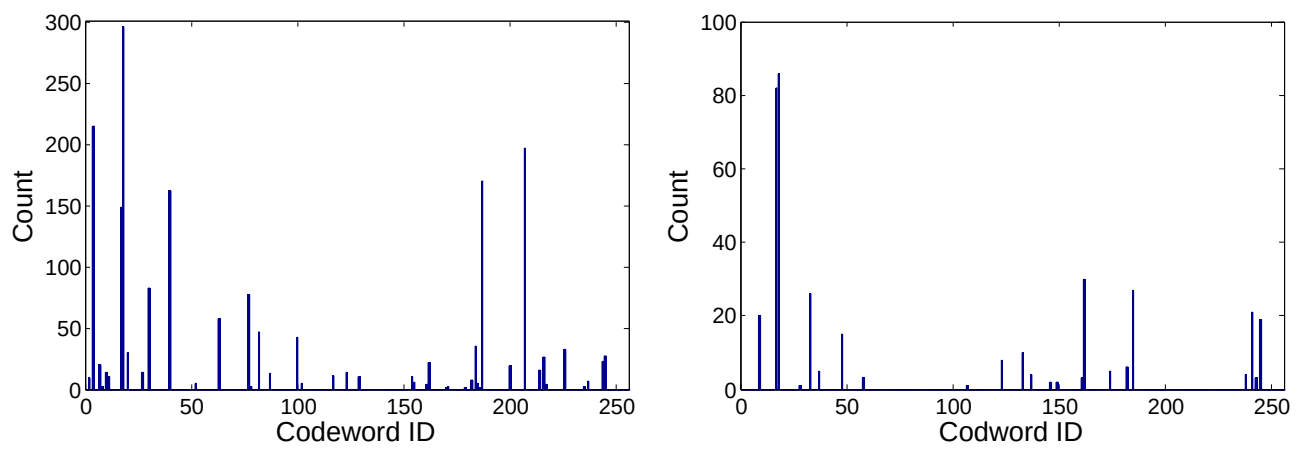

Fig. 3. Histogram of the number of frames or images that contribute to each codeword in the FV. The plot on the left is constructed using web images returned for a query and the one on the right is for a video in the TRECVID MED13 dataset.

To avoid these problems, we simply average the Fisher Vectors of the top ranked videos to obtain a representation of the query for pseudo-relevance. We then calculate cosine similarity with this mean vector to obtain the final ranked list. This is very efficient and robust to outliers due to averaging. Importantly, this re-ranking step replaces the initial web image based representation of the query with a video representation obtained from the same domain as the target videos, thus reducing the effect of domain shift in the similarity computation.

## IV. Fast Fisher Vector Products

Similarity computations employ inner products between Fisher Vectors are computationally expensive due to high feature dimensionality. We speed up these operations through a Fast Fisher Vector Product (FFP) algorithm which takes advantage of the sparsity of Fisher Vectors to accelerate the inner product procedure without loss in accuracy.

### A. Sparsity in FVs of Videos and web images

Fisher Vectors constructed from SIFT features extracted from images are dense (around 50% of the entries are non-zero) [43] due to the significant appearance variation of small patches within an image. However, due to temporal continuity between frames, CNN features of video frames are quite similar to each other. Moreover, search results returned by web image search can contain very similar image subsets. Hence, in both cases, many images contribute to the same set of codewords, resulting in sparse Fisher Vectors (we find that only 10% of the entries are non-zero when $K$ is large). The histogram in Fig. 3 illustrates the number of frames or images that contribute to each codeword ($K = 256$) in two Fisher Vectors (one constructed from a web image collection and the other from the frames of one video). Although in theory each sample (video frame/web image) contributes to every codeword in a Fisher Vector, the probabilities for most codewords are so low that they fall below measurable precision. Fig. 3 might suggest that we can reduce feature dimensionality by reducing the number of codewords, since most of them have low probabilities for any particular sample. However, Fig. 2 shows that reducing the number of codewords leads to a consistent decrease in performance—this is likely because different samples have different sparsity patterns and each codeword is useful for some data.

Therefore, we need to accelerate the inner product without reducing the feature dimensionality. A simple way to achieve this would use a sparse matrix representation, which only stores the non-zero elements in the matrix and their indices. In this case, a nonzero entry arises from a non-zero codeword, which has a nonzero contribution from at least one sample. If we multiply a sparse matrix with a column vector, we theoretically expect to observe a linear speedup with respect to the number of zeros in the matrix. However, this operation only takes advantage of sparsity in the matrix or the vector, but not both. It would still need to multiply some non-zero elements with zeros in the vector or vice-versa [57].

A sparse representation would also require less memory to store the FVs. For example, if 90% of the codewords of the FV are zero, it would lead to reduction in memory required by a factor of 10. For a general matrix, we would need to store all indices of the non-zero entries, but this is not required for FVs. This is because zeros are not randomly distributed in a FV, but instead have a grouping structure. If no video frame or web image $x_t$ contributes to a codeword $i$ (i.e. the codeword has an extremely low probability which falls below measurable precision for all samples), then all values for that codeword $\mathcal{G}_{\mu_i}, \mathcal{G}_{\sigma_i}$ are zero. This trick was proposed in [43] for compressing sparse FVs. We will show how this property can also be used to accelerate the computation of inner products between sparse FVs.

### B. Lossless Matching Algorithm

Let $Q, V_i \in \mathbf{R}^N$ denote the FV corresponding to web images and video frames respectively. Let $T$ denote the database of videos such that, $T = \{V_1, V_2, ..., V_M\}$, where $M$ is the number of videos. Assume the FV is constructed from a GMM of $K$ codewords, where each codeword is of dimension $D$. Since the FV includes differences from the mean and the variance for each codeword, $N = 2DK$. Compute $I^T = \{I_1^T, I_2^T, .., I_j^T, .., I_M^T\}$, where, $I_j^T \subseteq \{1, 2, ..., K\}$ denotes the set of non-zero codeword indices in $V_j$. Let $I^Q$ denote the set of non-zero codewords indices in the Fisher Vector for query $Q$. Once the query is provided, $I^Q$ is constructed by checking if the sub-vector corresponding to each codeword is non-zero. Compute the set intersection $S^T = \{S_1^T, S_2^T, .., S_j^T, .., S_M^T\}$, where, $S_j^T = I_j^T \cap I^Q$. Finally, inner product $P_j$ between $Q$ and $V_j$ is computed, only for codewords in $S_j^T$.

### C. Analysis

First, we assume that each codeword across the video database has equal probability $p^T$ of being a non-zero codeword. We also assume that each codeword in the query has equal probability $p^Q$ of being a non-zero codeword. Therefore, the probability that a codeword contributes a non-zero value to the inner product is $p^T p^Q$, and the expected number of operations for performing an inner product between a video and a query Fisher Vector is $2KDp^T p^Q$. The intersection only needs to be computed between codeword indices ($I^T$ and $I^Q$). So, after obtaining $I_j^T$ and $I^Q$, it only requires $\min(|I_j^T|, |I^Q|)$ operations to construct $S_j^T$ (using a hashtable for the codeword indices). Thus, the expected number of operations for calculating $S_j^T$ is $K \min(p^T, p^Q)$. Therefore, the expected number of operations required for matching in VRFP is $K \min(p^T, p^Q) + 2KDp^T p^Q$. The expected speedup over the brute force inner product algorithm is given by:

$$ES_{unbiased} = \frac{2KD}{2KDp^T p^Q + K \min(p^T, p^Q)} = \frac{1}{p^T p^Q + \frac{\min(p^T, p^Q)}{2D}} \quad (5)$$

If $C_1$ elements of $T$ are non-zero and $C_2$ elements of $Q$ are non-zero, then $p^T$ and $p^Q$ can be approximated as, $p^T = \frac{C_1}{MN}$ and $p^Q = \frac{C_2}{N}$ respectively. This shows that the number of arithmetic operations required by the algorithm is quadratic in the sparsity of the Fisher Vectors (assuming intersection computation time is bounded by a constant due to large D, D = 256 in our case). Note that sparsity is defined as the proportion of non-zeros in a matrix. In the case of sparse matrix multiplication, intersection computation is not efficient, as in the general case the grouping structure present in Fisher Vectors is absent. Therefore, intersection computation would take $2KD \min(p^T, p^Q)$ operations. Consequently, the number of arithmetic operations required for performing sparse multiplication would be linear in the sparsity of the Fisher Vectors. The time complexity for different algorithms is shown in Table I. One could view FFP as creating an index of non-zero codewords in a Fisher Vector and then computing inner product only within the index. This is essentially equivalent to building an inverted index of non-zero codewords in the Fisher Vectors.

This analysis is valid when all codewords have equal probability of being non-zero. However, few codewords may be non-zero most of the time. In this case, the previous analysis will not hold. Suppose $K - X$ ($X << K$) codewords are equally probable to be non-zero with a low probability $p_l$, while $X$ of them are equally probable to be non-zero with a high probability $p_h$. For simplicity, assume that the sparsity patterns of web images and video frames are similar, i.e., $p_l^T = p_l^Q$ and $p_h^T = p_h^Q$. The expected speedup per codeword (ignoring the intersection computation time) in this case would be:

$$ES_{biased} = \frac{K}{Xp_h^2 + (K - X)p_l^2} \quad (6)$$

TABLE I
MATRIX MULTIPLICATION (MM) VS FAST FV PRODUCTS

| **Method** | Naive MM | Sparse MM | FFP (Ours) |
|---|---|---|---|
| Complexity | $\mathcal{O}(N)$ | $\mathcal{O}(min(p^Q, p^T)N)$ | $\mathcal{O}(p^Q p^T N)$ |

### D. Lossy Matching Algorithm

Generally, $p_l$ is around 0.1 (10% of the elements are non-zero), therefore $p_l^2$ is very small. However, even if $p_h$ is 0.5, it would significantly slow down the algorithm. Therefore, a simple trick to speed up the algorithm would be to remove the codewords which are non-zero with a very high probability. As we will show in our experiments, with 256 codewords, removing only 2 codewords can yield a speedup of 20% without a significant loss in accuracy. In the case of normal multiplication, removing 2 codewords would have resulted in a speedup of less than 1%.

## V. EXPERIMENTAL RESULTS

### A. Datasets

We evaluate our method on three event detection datasets.

**TRECVID MED13/14 [1]**. The TRECVID MED 2013/2014 dataset consist of unconstrained Internet videos collected by the Linguistic Data Consortium from various Internet video web sites. Each video contains only one complex event or it is a background video and does not contain content related to any of the query events. There are in total 30 complex events in this dataset: “E1: birthday party”, “E2: changing a vehicle tire”, “E3: flash mob gathering”, “E4: getting a vehicle unstuck”, “E5: grooming an animal”,“E6: making a sandwich”, “E7: parade”, “E8: parkour”, “E9: repairing an appliance”, “E10: working on a sewing project”, “E11: attempting a bike trick”, “E12: cleaning an appliance”, “E13: dog show”, “E14: giving directions to a location”, “E15: marriage proposal”, “E16: renovating a home”, “E17: rock climbing”, “E18: town hall meeting”, “E19: winning a race without a vehicle”, “E20: working on a metal crafts project”, “E21: bee keeping”,“E22: wedding shower ”, “E23: non-motorized vehicle repair”, ‘E24: fixing musical instrument”, “E25: horse riding”, “E26: fellling a tree”, “E27: parking a vehicle”, “E28: playing fetch”, “E29: tailgating”, “E30 tuning musical instrument”. Videos of the first 20 events together with background videos (around 23,000 videos), form a test set of 25,000 videos for the MED13 testset, and the last 20 events with similar background videos form the MED14 testset.

We evaluate using the EK0 setting on the TRECVID dataset. EK0 is a setting for the TRECVID dataset, in which no training videos are provided for the event query. An event definition is provided as a single sentence. Further, a detailed description is available for the query in the form of a paragraph. Common objects, scenes and actions occurring in the event are also listed. Zero-shot learning based algorithms rely on this detailed description (to perform semantic matching with pre-trained concepts) for performing retrieval. The EK0 setting also allows for the manual selection of concepts at test time. The algorithm could obtain an initial concept list using semantic matching (like using word2vec) and final concept selection could be manual to improve performance. However, a detailed description is not usually provided by a typical user querying a search engine nor is manual selection of concepts a reasonable assumption in practice. Hence, VRFP does not utilize the detailed description provided in the EK0 setting but only uses the event name. VRFP is completely automatic, so no manual intervention is needed.

**Columbia Consumer Videos (CCV) [29]**. The CCV dataset contains 9,317 videos collected from YouTube with annotations of 20 semantic categories: "E1: basketball", "E2: baseball", "E3: soccer", "E4: ice skating", "E5: skiing", "E6: swimming", "E7: biking", "E8: cat", "E9: dog", "E10: bird", "E11: graduation", "E12: birthday","E13: wedding reception", "E14: wedding ceremony", "E15: wedding dance", "E16: music performance", "E17: non-music performance", "E18: parade", "E19: beach", "E20: playground". This dataset is evenly split into 4,659 training videos and 4,658 dataset videos. We focus on the scenario where no training videos are available, so we only run our method on the dataset videos and calculate the mAP.

### B. Implementation Details

For each category in the dataset, we use the name of the query to search for and download all images from Google. After obtaining web images, we build our query representation. On average, around 700 images are downloaded for each event. We sample one frame every 2 seconds for TRECVID MED and CCV dataset. Using the implementation of AlexNet [30] provided in Caffe [23], we extract 4,096 dimensional fc7 layer features for web images and video frames.

To compute Fisher Vectors, we first reduce the original features to 256 dimension using principal component analysis (PCA), and then use 256 components for Fisher Vectors. For VLAD, the CNN features are also reduced to 256 dimensions and the number of clusters is set to 256. The TRECVID dataset contains 4,992 background videos, which do not contain any test event. These background videos are used to learn the PCA projection, Fisher Vector GMM components and cluster centers for VLAD. Finally, VLFeat [49] is used to generate VLAD and Fisher Vectors.

We also apply another square root normalization on the Fisher Vector built from video frames (when comparing similarity between web images and video frames). This is because correlation between web images is less compared to video frames. Thus, values in the Fisher Vector for video frames have a more peaky structure compared to web images. Therefore, we apply square-root normalization (which normalizes the peaks) only to videos but not web images. To boost sparsity, we shrink values whose magnitude falls below $10^{-3}$ in the FV to zero. We use the top 50 ranked videos in the initial ranked list for re-ranking. When SVM is used for re-ranking, the bottom 1,000 videos are used as negatives. Liblinear [14] is used to train and test all these linear SVMs, and the $C$ parameter is set to 1 for all SVMs.

In total, it takes less than a second for VRFP to perform retrieval for 100,000 videos: downloading images from the web takes less than 500 milliseconds, CNN features on 700 images are extracted in 150 milliseconds (using AlexNet implementation of Torch on Titan X), building Fisher Vector of web images takes 30ms, and feature matching and re-ranking is performed in less than 240 milliseconds when using 130,072 dimensional Fisher Vectors. A compressed Fisher Vector representation (using sparsity) occupies only 1GB memory for 100,000 videos (with 16 bit floating point precision). All performance experiments were carried out on an Intel Ivy Bridge E5-2680v2 processor using only a single core.

For fairness, when comparing running time of all methods, we use double precision for non-integer storage (as MATLAB sparse matrix multiplication requires a double input). Other than sparse matrix multiplication, every baseline including Product Quantization, High Variance Subspaces First, naive matrix multiplication and the proposed fast FV product code was written in C++. Each code was optimized taking cache locality into account. Since our method may produce different intersections for each query, the computation time varies per query. Therefore, we average the time required for all 20 queries in TRECVID MED13 in our results.

### C. Performance of Different Representations for Videos and Images

We compare different image and video representations in Table II using mAP. For a fair comparison, these results do not include the re-ranking step. Wherever SVM is not mentioned (i.e., Max Pooling, Avg Pooling, VLAD and Fisher Vectors in Table II), we first generate the representations of web images and dataset videos using the corresponding method, and then use cosine similarity to measure distance between the representations of web images and dataset videos. Avg + SVM trains an SVM using the web images as positives and 1,000 randomly selected web images of other queries as negatives. Then, we apply SVM on average pooled video representations to rank the videos. Training an SVM on a large number of samples can take a significant amount of time. It takes 300ms to train a linear SVM, while building FV only requires 30ms. However, we provide this comparison for completeness.

We make the following observations from Table II: 1) SVM, VLAD, and Fisher Vector perform better than average pooling and max pooling in all cases; 2) Fisher Vector perform better than VLAD, because it can model second order statistics; 3) Max pooling gives the lowest mAP because it is very sensitive to noise and not suitable for matching noisy web images with video frames; 4) Since TRECVID MED13/14 are complex event detection tasks and contain diverse videos, using a VLAD or Fisher Vector representation for both web images and videos yields good performance for matching. However, on the CCV dataset, SVM using average pooled features performs better. This is because the CCV dataset contains videos with only a few frames that are all very similar. Since CCV contains very simple events, the retrieved web images also have similar appearance. Nevertheless, in section V-E, we will show that the video retrieval performance of Fisher Vector outperforms SVM on the CCV dataset after re-ranking.

TABLE II
COMPARISON AMONG DIFFERENT REPRESENTATIONS

| Method | MED13 | MED14 | CCV |
|---|---|---|---|
| Max Pooling | 2.01% | 0.78% | 7.71% |
| Avg Pooling | 7.78% | 2.68% | 28.81% |
| Avg + SVM | 9.17% | 6.0% | **37.28%** |
| VLAD | 11.36% | 6.60% | 28.93% |
| Fisher Vectors | **14.1%** | **8.49%** | 32.65% |

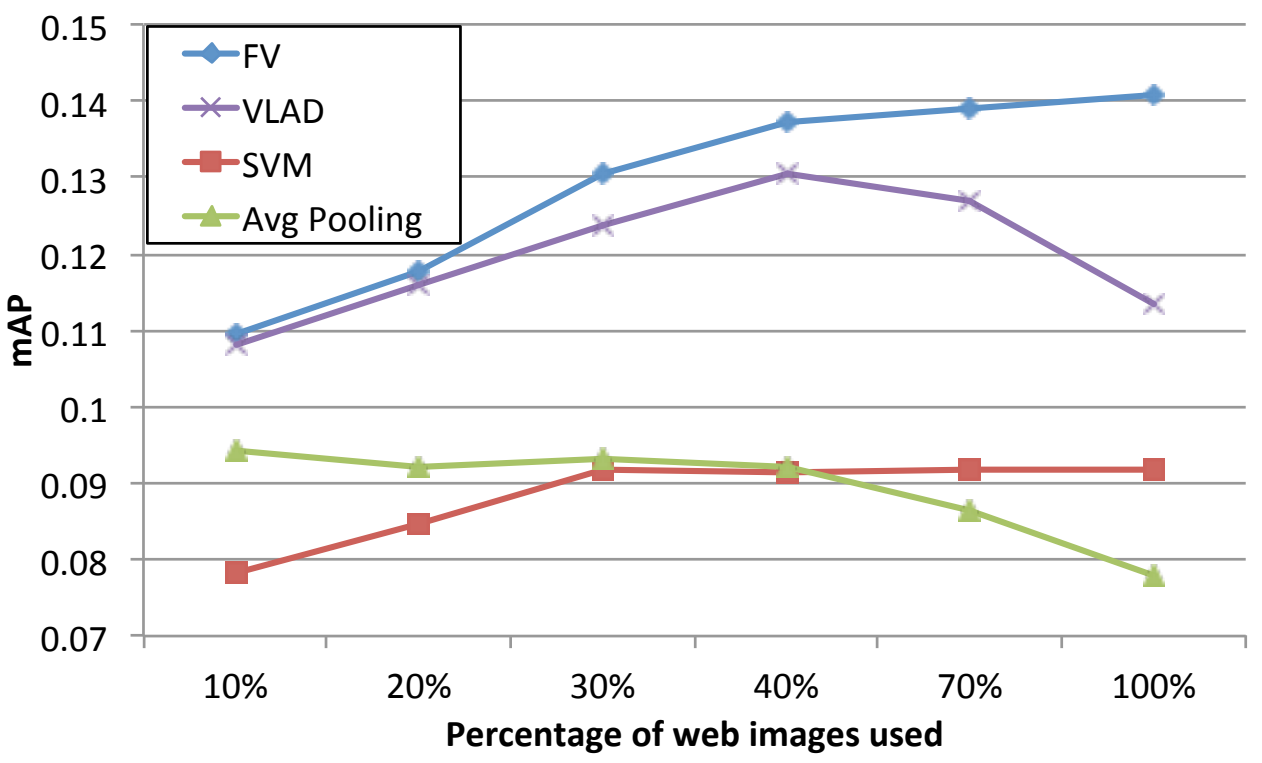

Fig. 4. Comparison of representations on TRECVID MED13 as the number of web images is varied. Around 700 images are downloaded for each event.

### D. Robustness Analysis of Representations

We conducted experiments on the TRECVID MED13 dataset to show the advantages of the Fisher Vector representation compared to VLAD, average pooling and SVM for matching noisy web image collection with video frames. Max pooling is excluded due to its poor performance. We investigate the following key factors that may affect retrieval accuracy and evaluate the robustness of different representations:

**Number of Web Images** For each event, we download around 700 images. In Fig. 4, we vary the number of web images for building the representations and training SVM, i.e., only the top 10%, 20%, 30%, 40%, 70%, and 100% retrieved images from the search engine are used. We show the mAP on MED13 test dataset for each case.

Fig. 4 shows that the performance of FV, VLAD and SVM improves significantly until the top 40% web images are used. When more web images are added, the mAP of SVM does not improve and even drops as the number of web images approaches 100%. This is because the top ranked results returned by an image search engine contain fewer noisy images, while the lower ranked images are often more noisy and sometimes completely irrelevant to the query. Thus, using these noisy images as positive samples to train the SVM hurts performance. This problem is more severe for VLAD encoding and average pooling. For average pooling, its mAP drops consistently when more images are used to calculate the average. This is because the top 10% of the web images share similar appearance, and average pooling of these images serves as a nearest mean classifier to these images. However, when more images are used, diverse results (including outliers) affect the mean estimation and hurt performance. The lower-ranked noisy web images also significantly hurt the performance of VLAD encoding. We attribute this to the fact that VLAD uses a hard assignment instead of soft assignment used in FV, which make it more sensitive to noisy samples.

In contrast to SVM, VLAD and average pooling, the FV representation always benefits from more training images even when some of them are noisy. This means that FV is a better choice when calculating similarity between a noisy web image collection and frames of a video.

**Outliers in Web Images** Outliers in image search results are known to cause problems when training models on web images. Among the downloaded web images, we observe around 10% outliers and lower-ranked images tend to contain more outliers. Some works [4], [7] show that by identifying and eliminating outliers, the web-supervised classifier yields higher accuracy for image/video retrieval. In this paper, we employ state-of-the-art outlier removal algorithms based on an autoencoder [52]. We train an autoencoder only using the web images of a query. Formally, for web images of a query whose feature vectors are $\{x_1, x_2, ..., x_n\}$, an autoencoder minimizes the sum of reconstruction error:

$$\mathcal{J}(f) = \sum_{i}^{n} \epsilon_i = \sum_{i}^{n} \|f(x_i) - x_i\|^2 \tag{7}$$

where $f(\cdot)$ is a neural network with a single hidden layer, and $\epsilon_i$ denotes the squared loss of sample $x_i$. The main idea is that if we train an autoencoder using the web images for a query, the positive samples (relevant images of the query) have a smaller reconstruction error than outliers (irrelevant images of the query). Thus, if we apply 2-means clustering on image reconstruction errors, the outliers would be in the cluster with a higher reconstruction error. The outlier removal algorithm improves the mAP for all representations as shown in Table. III. For FV, the outlier removal only gives a marginal improvement, which means the FV is more robust to the outliers in web image search. Thus, FV is more suitable for building representations for noisy data like web images, and an outlier removal algorithm is not necessary to achieve high accuracy. Therefore, we do not include this outlier removal method in FV-based final results. This also avoids the need to train an autoencoder after the web images are obtained, which hurts real-time performance.

**Noisy Samples** To further illustrate the superiority of FV representation, we evaluate FV and the other two representations by adding 10% to 50% negative images to the positive training data. More specifically, for SVM, we sample some images in the negative training set and change their labels to positive and re-train the SVM; for FV, VLAD and average pooling, these negatives, together with the original positives, are used to build the feature vectors and averaging. The results in Fig. 5 suggests that when more and more negative images are used, the performance drop of FV is less than SVM, while average pooling and VLAD suffer more from negative samples.

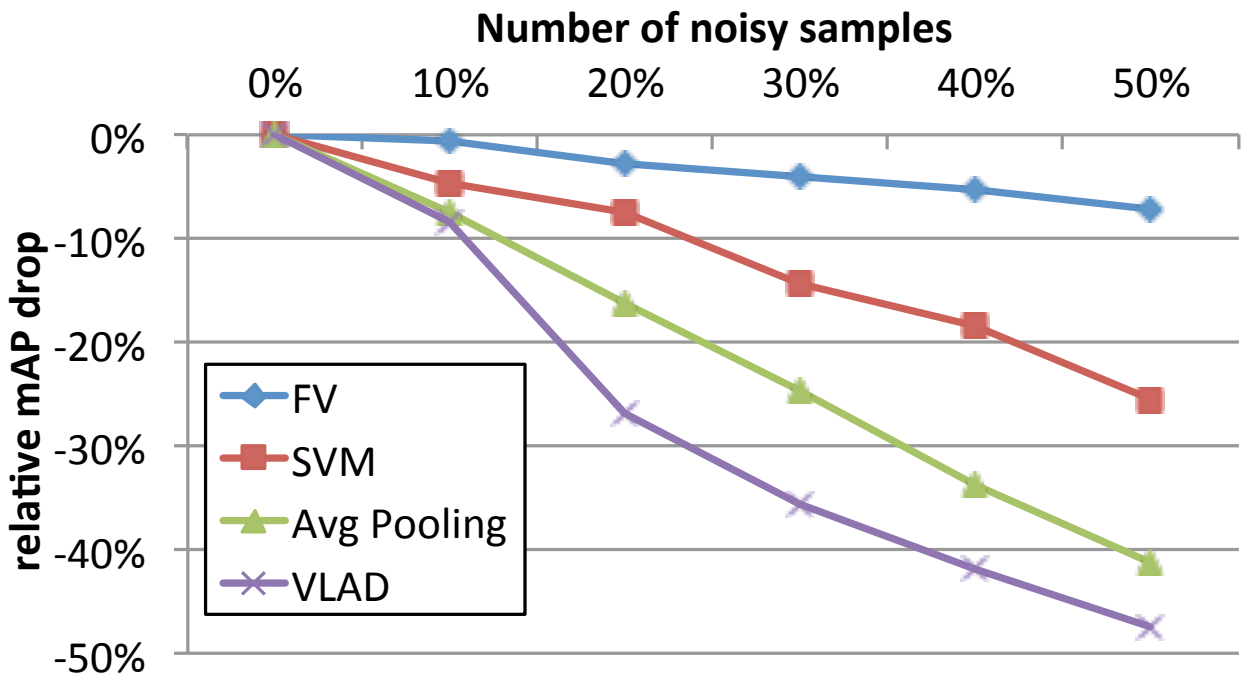

Fig. 5. Relative mAP drop of FV, SVM and average pooling when gradually adding negative images from other queries.

TABLE III
MAP IMPROVEMENT FOR MED13 DATASET BY OUTLIER REMOVAL

| Method | Avg Pooling | SVM | VLAD | FV |
|---|---|---|---|---|
| with outliers | 7.78% | 9.17% | 11.36% | 14.10% |
| without outliers | 8.60% | 9.92% | 12.83% | 14.25% |
| Relative change | 10.54% | 8.18% | 12.94% | 1.06% |

Consequently, we conclude that FV is a better choice than VLAD, SVM and average pooling for on-the-fly video retrieval using web images.

### *E. Exploring Re-ranking Methods*

We show the effectiveness of re-ranking and compare three different re-ranking strategies in Table IV. Avg Pooling + SVM trains a linear SVM classifier on average pooled video features using top-ranked videos as positives and bottom ranked videos as negatives, and re-ranks using the classifier response. Similarly, FV + SVM uses the FV of videos to train an SVM and performs re-ranking. These two SVM based methods use pseudo relevance based re-ranking as in [45]. From Table IV, we can see the most efficient and best performing method for this task is computing an average Fisher Vector for top ranked videos and then computing its cosine similarity with the dataset videos (FV Similarity). While FV + SVM can achieve similar performance as FV similarity, it takes more than 2 seconds to train an SVM due to high dimensional features.

TABLE IV
COMPARISON AMONG DIFFERENT REPRESENTATIONS

| Method | MED13 | MED14 | CCV |
|---|---|---|---|
| FV (No re-ranking) | 14.10% | 8.49% | 32.65% |
| Avg Pooling + SVM | 11.95% | 6.28% | 36.29% |
| FV + SVM | 15.91% | 8.72% | 38.08% |
| FV similarity | **16.44%** | **9.67%** | **40.81%** |

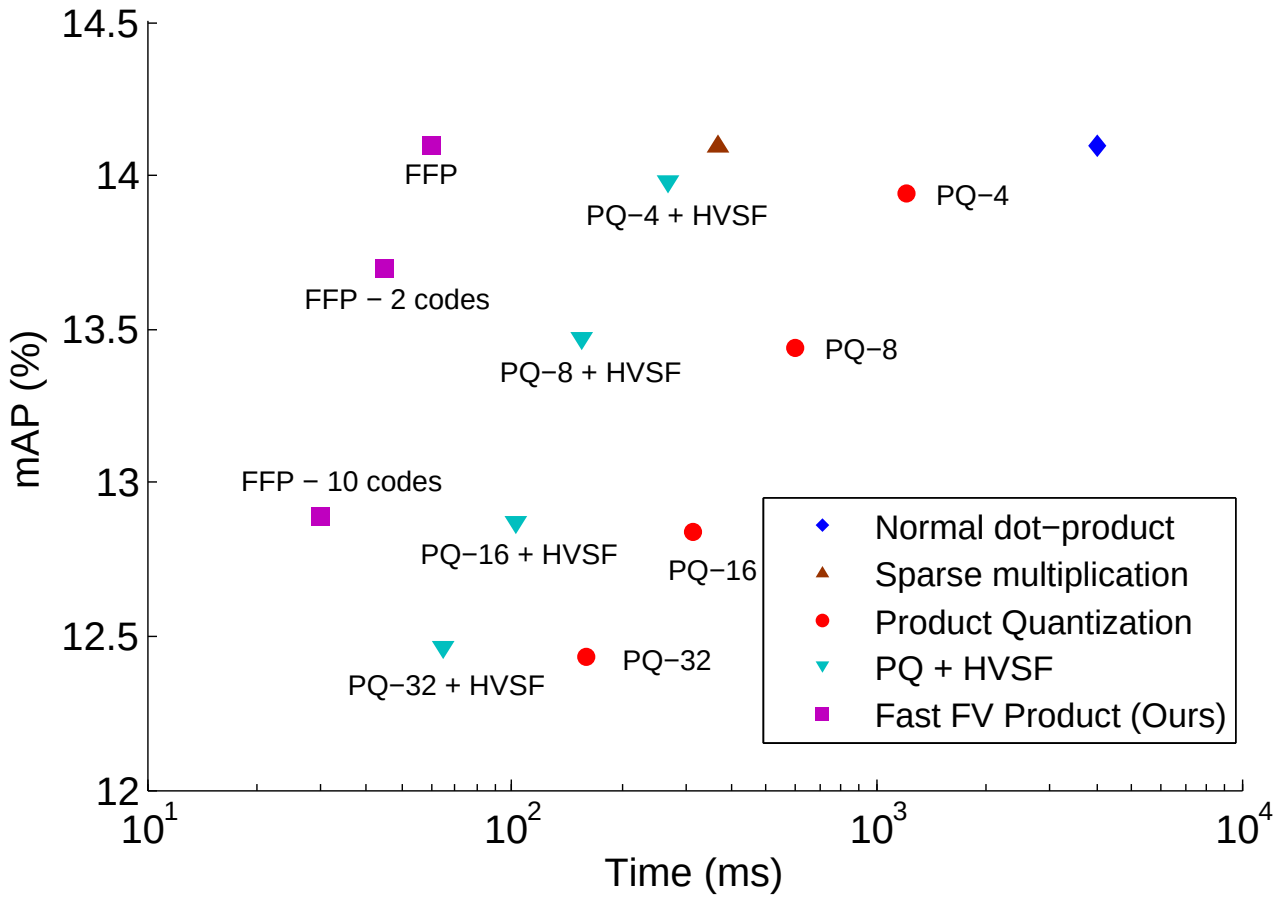

Fig. 6. Performance (measured by mAP) vs. matching time on TRECVID MED13 dataset. *PQ-k* denotes Product Quantization with subvector length equal to $k$. HVSF denotes the High Variance Space First algorithm described in [56]. *FFP - k codes* is our fast FV product method with $k$ codewords removed.

Avg Pooling + SVM does not work as well as Fisher Vectors since features of different videos can be very different and averaging the features further leads to a loss of information. Some other re-ranking methods like [24], [25] take around 2.5 minutes to run. Thus, methods like SVM are not well suited for the re-ranking step.

For the CCV dataset, re-ranking the initial results generated by FV cosine similarity improves mAP to 40.81%. However, if we use the same re-ranking for Avg + SVM method (in Table II), the mAP only improves from 37.28% to 38.73%, which is lower than the re-ranking result of FV (re-ranking improve FV from 32.65% to 40.81%). The videos in the initial top ranked list obtained using FV are diverse in appearance, so the average FV of top-ranked videos contains more useful information. Thus, FV is generally a good choice to match web images and video frames and to re-rank the dataset videos.

### *F. Speedup using Fast Fisher Vector Products*

As shown in Fig. 3, the Fisher Vectors of web images and video frames are sparse. For web images, we find that only 15% of the codewords are non-zero in all 20 queries for TRECVID MED13. For video Fisher Vectors, only 7% of the codewords are non-zero. Therefore, just using sparse matrix multiplication, we obtain a speedup of more than 11 (360ms vs 4000ms) over naive matrix multiplication (NMM). When we use the proposed Fast Fisher Vector products, we obtain an additional 6x speedup over sparse matrix multiplication (60ms vs 360ms, 66x faster than NMM). This is because as we increase sparsity linearly, our computation reduces quadratically. For sparse matrix multiplication, computation only decreases linearly with sparsity. Surprisingly, even our lossless version of Fast Fisher Vector Products is 2x faster than computing a naive inner product between 4096 dimension CNN features. Using the lossy algorithm, we obtain a speedup of 2x (30ms, 132x NMM) over the lossless algorithm just by removing 10 codewords ($<$ 4% of whole codeword size). This is because high probability codewords dominate the sum in the denominator of Equation 6.

TABLE V
AP FOR EACH EVENT IN TRECVID MED13 DATASET

| **MED13** | E1 | E2 | E3 | E4 | E5 | E6 | E7 | E8 | E9 | E10 | E11 | E12 | E13 | E14 | E15 | E16 | E17 | E18 | E19 | E20 | mAP |
|---|---|---|---|---|---|---|---|---|---|---|---|---|---|---|---|---|---|---|---|---|---|
| EventNet [54] | 9.7 | 32 | 0.3 | 1 | 1.8 | 5.7 | 27.4 | 18.1 | 4.3 | 0.9 | 0.8 | 2.9 | **47.0** | 0.1 | 0.5 | 0.3 | **7.5** | **16.1** | 0.1 | 0.5 | 8.86 |
| Pair-Concept [45] | **17.2** | 43.8 | 42.2 | **50.3** | 6.43 | 12.6 | 15.8 | 6.56 | 13.7 | 3.36 | 8.14 | **7.94** | 1.11 | 0.36 | 0.26 | **3.13** | 1.73 | 1.32 | 0.13 | 1.02 | 11.8 |
| Concept Proto [33] | 15.4 | 32 | 27.1 | 40.6 | 9.5 | 16.4 | 24 | 11.2 | 21.3 | 8.9 | 6.1 | 2.6 | 1.1 | 0.8 | 0.5 | 2.6 | 3.6 | 3.5 | **10.1** | **1.4** | 11.9 |
| TagBook [34] | 15.5 | 33.7 | 17.4 | 31.2 | **20.1** | 9.9 | 18.5 | **21.5** | 21.1 | **9.8** | 6.6 | 2.3 | 20.0 | 0.5 | 0.3 | 1.8 | 2.6 | 14.8 | 9.9 | 0.2 | 12.9 |
| VRFP web-only (Ours) | 9.4 | 42.7 | 38.5 | 42.4 | 7.6 | 13.7 | 26.9 | 15.5 | 17.0 | 3.7 | 9.2 | 1.6 | 33.7 | **1.4** | 0.1 | 2.7 | 4.9 | 8.4 | 0.8 | 1.1 | 14.1 |
| VRFP re-ranking(Ours) | 10.8 | **44.4** | **42.7** | 50.2 | 7.5 | **19.3** | **30.1** | 11.4 | **31.5** | 2.2 | **17.2** | 1.9 | 45.3 | 0.6 | **0.9** | 1.4 | 7.4 | 1.5 | 1.6 | 0.6 | **16.4** |

TABLE VI
AP FOR EACH EVENT IN TRECVID MED14 DATASET

| **MED14** | E11 | E12 | E13 | E14 | E15 | E16 | E17 | E18 | E19 | E20 | E21 | E22 | E23 | E24 | E25 | E26 | E27 | E28 | E29 | E30 | mAP |
|---|---|---|---|---|---|---|---|---|---|---|---|---|---|---|---|---|---|---|---|---|---|
| TagBook [34] | 7.5 | 8.0 | 15.7 | 0.6 | 0.5 | **4.7** | 2.0 | **12.0** | 6.3 | 0.5 | 0.9 | **3.5** | **26.5** | 0.9 | 11.8 | **7.2** | 3.5 | **3.5** | 0.6 | 0.9 | 5.9 |
| AutoVisual [27] | 5.4 | **15.0** | **42.4** | **2.5** | **4.1** | 0.5 | **11.9** | 1.9 | **6.6** | **6.4** | 45.65 | 3.0 | 1.3 | 1.8 | **13.24** | 0.5 | 4.0 | 0.2 | 0.1 | 1.1 | 8.4 |
| VRFP web-only (Ours) | 9.4 | 2.6 | 29.2 | 0.7 | 1.1 | 2.7 | 5.0 | 9.4 | 1.1 | 1.1 | **70.2** | 1.0 | 3.6 | 6.3 | 6.6 | 1.9 | 7.9 | 2.4 | **2.0** | 5.8 | 8.5 |
| VRFP re-ranking (Ours) | **17.6** | 3.1 | 39 | 0.4 | 1.1 | 1.5 | 7.4 | 3.5 | 2.3 | 0.5 | 65.47 | 0.4 | 4.9 | **15.7** | 7.4 | 2.0 | **12.0** | 1.3 | 1.5 | **6.4** | **9.7** |

TABLE VII
AP FOR EACH EVENT IN CCV DATASET

| **CCV** | E1 | E2 | E3 | E4 | E5 | E6 | E7 | E8 | E9 | E10 | E11 | E12 | E13 | E14 | E15 | E16 | E17 | E18 | E19 | E20 | mAP |
|---|---|---|---|---|---|---|---|---|---|---|---|---|---|---|---|---|---|---|---|---|---|
| Large Concept [9] | 5.8 | 7.9 | 12.5 | 7.4 | 25.4 | 21.5 | 11.3 | 8.8 | 18.7 | 15.6 | 24.2 | 10.7 | **29.3** | 22.1 | 22.7 | 21.4 | 16 | 21.9 | 47 | 10.1 | 18 |
| EventNet [54] | **59.5** | **55** | **57.5** | **77.7** | 69.1 | 73.6 | 19.6 | 41.9 | 29 | 10.3 | 15.1 | **13.1** | 12.9 | **27** | 18.4 | 28.7 | **35.6** | **47.4** | 16.4 | 3.89 | 35.6 |
| VRFP web-only (Ours) | 11.0 | 38.1 | 34.8 | 41.8 | 57.9 | 73.4 | 49.0 | 37.7 | 27.2 | **22.5** | 32.0 | 9.82 | 7.02 | 9.84 | 36.0 | **31.0** | 12.0 | 37.4 | 55.6 | 28.9 | 32.7 |
| VRFP re-ranking (Ours) | 34.3 | 53.4 | 54.7 | 50.2 | **75.4** | **80.1** | **65.2** | **49.0** | **42.3** | 19.3 | **32.1** | 9.69 | 8.26 | 8.66 | **49.42** | 24.4 | 15.0 | 31.2 | **69.3** | **44.1** | **40.8** |

We also compare our algorithm with other state-of-the-art algorithms like product quantization (PQ) [20] and High Variance Subspaces First (HVSF) [56] which are used for fast retrieval of high dimensional vectors. We show results for product quantization and HVSF with 4 different sub-vector lengths. In HVSF, we select the top 20% subspaces with higher variances to calculate the initial ranked list and compute the extract inner product for top 500 videos [56]. Accuracy of PQ and HVSF drops when we increase the sub-vector length, although they do provide linear speedup. Fig. 6 compares the accuracy and efficiency of our system without re-ranking with these methods. Note that PQ and HVSF are approximate algorithms, which reduce accuracy. Our lossless algorithm obtains a 66x speedup over standard matrix multiplication without any loss in accuracy. If the inner product in the remaining codewords in $S^T$ needs to be accelerated, PQ can be applied on top of VRFP. To further reduce the elements in $S^T$, intersection can be computed with the subspaces in HVSF. The proposed algorithm is not an alternative to PQ or HVSF, but removes redundant multiplications which arise when computing inner products between sparse Fisher Vectors.

Fig. 7 further compares the query time and accuracy of our system with the methods without using Fisher Vectors as shown in TABLE II . Since Fisher Vector, VLAD, and Avg + SVM introduce different extra computational costs (e.g., generating FV, training SVM), different from Fig. 6 where we only show the matching time, we also include the extra computational time in this figure for a fair comparison. From this figure, we can see that our FFP method achieves the best performance with a short query time. Although the query time of Avg Pooling is similar to FFP, FFP has much higher mAP.

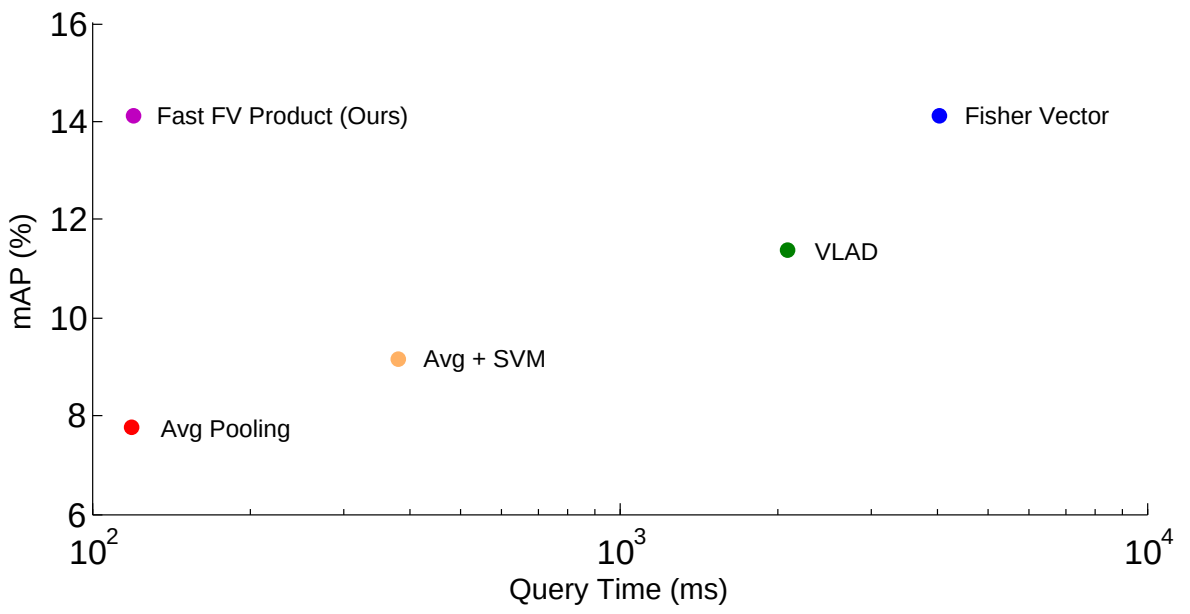

Fig. 7. Performance (measured by mAP) vs. query time of methods with normal inner product and our Fast FV product on TRECVID MED13 dataset.

TABLE VIII
COMPARATIVE RESULTS ON TRECVID MED13/14 AND CCV

| Method | MED13 | MED14 | CCV |
|---|---|---|---|
| Large Concept Bank [9] | 2.2% | - | 18.0% |
| Concept Discovery [7] | 2.3% | - | - |
| Object2Action [19] | 4.21% | - | - |
| Composite Concept [17] | 6.4% | - | - |
| AutoVisual [27] | 7.4% | 8.37% | - |
| EventNet [54] | 8.86% | - | 35.6% |
| MMPRF [25] | 10.1% | - | - |
| Pair-Concept [45] | 11.8% | - | - |
| Concept Prototypes [33] | 11.9% | - | - |
| Multi-Modal [50] | 12.6% | - | - |
| TagBook [34] | 12.9% | 5.9% | - |
| SPaR [24] | 12.9% | - | - |
| **VRFP(Ours)** | **16.4%** | **9.67%** | **40.8%** |

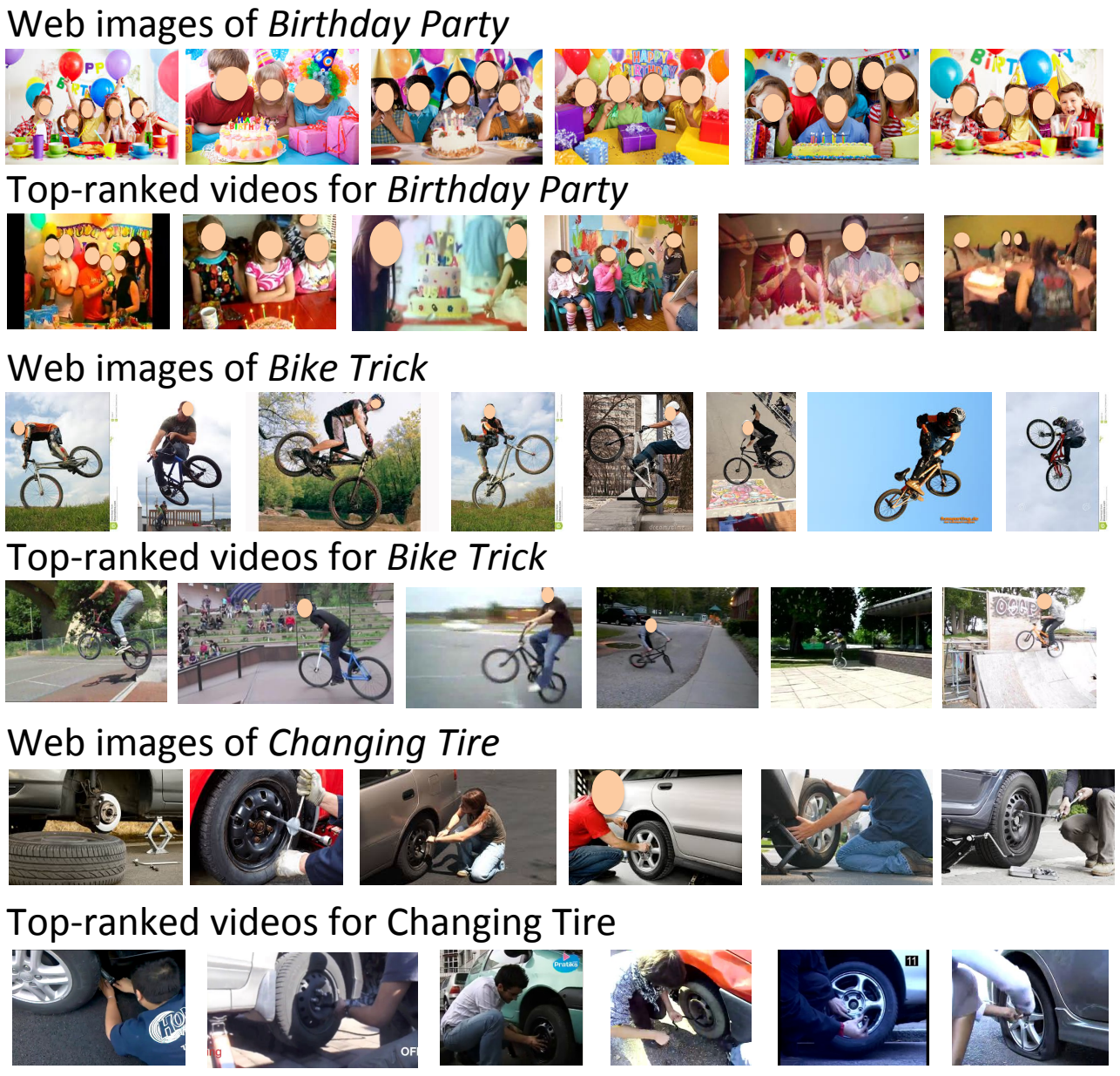

Fig. 8. Retrieved web images and top-ranked videos by VRFP for *Birthday Party*, *Bike Trick*, and *Changing Tire*.

### G. Comparison with Other Methods

We compare VRFP with other state-of-the-art methods that only use event names or descriptions to perform event retrieval. Table VIII shows the results of VRFP on TRECVID MED13/14 dataset and CCV datasets. Compared with other automatic methods, our method performs favourably. VRFP only requires a short query, instead of a long detailed description [25], [27], [33], [34], [50]. Methods like [19], [25], [27], [33], [34], [50] first build a large pre-defined concept bank. Then, relevant concepts are chosen based on their semantic similarities to the event description, and the selected concept detectors are used to rank database videos. Since the semantic gap is large for these pre-defined concept based approaches, their performance is lower when compared to VRFP. Although methods like [7], [45] use an image search engine to discover relevant concepts and use the downloaded web images to train concept detectors after the query is given, VRFP compares favourably to them because it uses a better representation for matching web images and video frames. Note that our pipeline is fully automatic and does not require any manual intervention. Methods like [25], [50] also use multi-modal features like automatic speech recognition (ASR), OCR or motion features like improved trajectories, etc. Note that for [27], we only report its automatic version instead of the one with manual inspection for fair comparison. Further, as shown in the previous section, our runtime performance is at least 10,000 times faster than the methods which leverage web data after the query is given, like [7], [45].

We also explore the possibility of using YouTube videos to perform on-the-fly retrieval. Downloading web videos of a query could take several seconds, so directly using web videos will hurt the real-time performance. Alternatively, for a query, we use 250 thumbnails downloaded from YouTube and build a Fisher Vector using both web images and video thumbnails. With the help of web video thumbnails, the mAP for TRECVID MED13 dataset improve from 16.44% to 16.83%, and mAP of MED14 improve from 9.67% to 9.99%. Since it would not be fair to use YouTube thumbnails for CCV dataset (as all videos in these datasets are from YouTube), we do not conduct this experiment on CCV.

In Table V, VI and VII we also show the AP scores of all events in TRECVID MED13/14 and CCV dataset, where VRFP with and without re-ranking are shown. These tables show that our method performs better and re-ranking improves results. Fig. 8 shows visual results of VRFP on three events from TRECVID MED13 dataset. It is clear that VRFP is effective in detecting complex events using web images.

## VI. Conclusion

We proposed VRFP, an on-the-fly video retrieval system that requires only a short text input query. Without any pre-defined concepts, VRFP retrieves web images for that query and builds a bag-based scalable representation via a Fisher Vector. VRFP uses simple matching and re-ranking strategies that do not require discriminative training to perform video retrieval. We showed that a Fisher Vector is robust to noise present in web images and also studied how it performs as the number of web images for training varies. To accelerate the computation of inner products between high dimensional Fisher Vectors, we presented a lossless algorithm in which the number of arithmetic operations decrease quadratically in terms of sparsity of Fisher Vectors. State-of-the-art results on three popular event retrieval datasets demonstrated the effectiveness of our approach.

## Acknowledgment

The authors would like to thank Yin Cui, Guangnan Ye, and Yitong Li for providing the AP for each event of their methods. The authors acknowledge the University of Maryland supercomputing resources http://www.it.umd.edu/hpcc made available for conducting the research reported in this paper.

## References

[1] Nist trecvid multimedia event detection (med) evaluation track, http://www.nist.gov/itl/iad/mig/med.cfm.
[2] R. Arandjelovic and A. Zisserman. Multiple queries for large scale specific object retrieval. In *BMVC*, pages 1–11, 2012.
[3] R. Arandjelovic and A. Zisserman. All about vlad. In *Computer Vision and Pattern Recognition (CVPR), 2013 IEEE Conference on*, pages 1578–1585. IEEE, 2013.
[4] K. Chatfield, R. Arandjelović, O. M. Parkhi, and A. Zisserman. On-the-fly learning for visual search of large-scale image and video datasets. *International Journal of Multimedia Information Retrieval*, 2015.
[5] K. Chatfield, K. Simonyan, and A. Zisserman. Efficient on-the-fly category retrieval using convnets and gpus. In *Asian Conference on Computer Vision*, pages 129–145. Springer, 2014.
[6] K. Chatfield and A. Zisserman. Visor: Towards on-the-fly large-scale object category retrieval. In *Asian Conference on Computer Vision*, Lecture Notes in Computer Science. Springer, 2012.
[7] J. Chen, Y. Cui, G. Ye, D. Liu, and S.-F. Chang. Event-driven semantic concept discovery by exploiting weakly tagged internet images. In *Proceedings of International Conference on Multimedia Retrieval*, page 1. ACM, 2014.
[8] X. Chen, A. Shrivastava, and A. Gupta. Neil: Extracting visual knowledge from web data. In *IEEE International Conference on Computer Vision (ICCV), 2013*.

[9] Y. Cui, D. Liu, J. Chen, and S.-F. Chang. Building a large concept bank for representing events in video. *arXiv preprint arXiv:1403.7591*, 2014.

[10] J. Dalton, J. Allan, and P. Mirajkar. Zero-shot video retrieval using content and concepts. In *Proceedings of the 22nd ACM international conference on Conference on information & knowledge management*. ACM, 2013.

[11] J. Deng, W. Dong, R. Socher, L.-J. Li, K. Li, and L. Fei-Fei. Imagenet: A large-scale hierarchical image database. In *Computer Vision and Pattern Recognition, 2009. CVPR 2009. IEEE Conference on*, pages 248–255. IEEE, 2009.

[12] S. K. Divvala, A. Farhadi, and C. Guestrin. Learning everything about anything: Webly-supervised visual concept learning. In *IEEE Conference on Computer Vision and Pattern Recognition (CVPR), 2014*.

[13] L. Duan, D. Xu, and S.-F. Chang. Exploiting web images for event recognition in consumer videos: A multiple source domain adaptation approach. In *IEEE Conference on Computer Vision and Pattern Recognition (CVPR), 2012*.

[14] R.-E. Fan, K.-W. Chang, C.-J. Hsieh, X.-R. Wang, and C.-J. Lin. Liblinear: A library for large linear classification. *The Journal of Machine Learning Research*, 9:1871–1874, 2008.

[15] R. Fergus, L. Fei-Fei, P. Perona, and A. Zisserman. Learning object categories from google's image search. In *Computer Vision, 2005. ICCV 2005. Tenth IEEE International Conference on*, volume 2, pages 1816–1823. IEEE, 2005.

[16] Y. Gong and S. Lazebnik. Iterative quantization: A procrustean approach to learning binary codes. In *Computer Vision and Pattern Recognition (CVPR), IEEE Conference on*, pages 817–824. IEEE, 2011.

[17] A. Habibian, T. Mensink, and C. G. Snoek. Composite concept discovery for zero-shot video event detection. In *Proceedings of International Conference on Multimedia Retrieval*, page 17. ACM, 2014.

[18] A. Habibian, T. Mensink, and C. G. Snoek. Videostory: A new multimedia embedding for few-example recognition and translation of events. In *Proceedings of the ACM International Conference on Multimedia*. ACM, 2014.

[19] M. Jain, J. C. van Gemert, T. Mensink, and C. G. Snoek. Objects2action: Classifying and localizing actions without any video example. In *Proceedings of the IEEE International Conference on Computer Vision*, pages 4588–4596, 2015.

[20] H. Jegou, M. Douze, and C. Schmid. Product quantization for nearest neighbor search. *Pattern Analysis and Machine Intelligence, IEEE Transactions on*, 33(1):117–128, 2011.

[21] H. Jégou, M. Douze, C. Schmid, and P. Pérez. Aggregating local descriptors into a compact image representation. In *Computer Vision and Pattern Recognition (CVPR), 2010 IEEE Conference on*, pages 3304–3311. IEEE, 2010.

[22] H. Jégou, F. Perronnin, M. Douze, J. Sanchez, P. Perez, and C. Schmid. Aggregating local image descriptors into compact codes. *Pattern Analysis and Machine Intelligence, IEEE Transactions on*, 34(9):1704–1716, 2012.

[23] Y. Jia, E. Shelhamer, J. Donahue, S. Karayev, J. Long, R. Girshick, S. Guadarrama, and T. Darrell. Caffe: Convolutional architecture for fast feature embedding. In *Proceedings of the 22nd ACM international conference on Multimedia*, pages 675–678. ACM, 2014.

[24] L. Jiang, D. Meng, T. Mitamura, and A. G. Hauptmann. Easy samples first: Self-paced reranking for zero-example multimedia search. In *ACM MM*, 2014.

[25] L. Jiang, T. Mitamura, S.-I. Yu, and A. G. Hauptmann. Zero-example event search using multimodal pseudo relevance feedback. In *Proceedings of International Conference on Multimedia Retrieval*. ACM, 2014.

[26] L. Jiang, S.-I. Yu, D. Meng, T. Mitamura, and A. G. Hauptmann. Bridging the ultimate semantic gap: A semantic search engine for internet videos. In *International Conference on Multimedia Retrieval*, 2015.

[27] L. Jiang, S.-I. Yu, D. Meng, Y. Yang, T. Mitamura, and A. G. Hauptmann. Fast and accurate content-based semantic search in 100m internet videos. In *Proceedings of the 23rd Annual ACM Conference on Multimedia Conference*, pages 49–58. ACM, 2015.

[28] Y.-G. Jiang, Q. Dai, T. Mei, Y. Rui, and S.-F. Chang. Super fast event recognition in internet videos. *IEEE Transactions on Multimedia*, 17(8):1174–1186, 2015.

[29] Y.-G. Jiang, G. Ye, S.-F. Chang, D. Ellis, and A. C. Loui. Consumer video understanding: A benchmark database and an evaluation of human and machine performance. In *Proceedings of the 1st ACM International Conference on Multimedia Retrieval*, page 29. ACM, 2011.

[30] A. Krizhevsky, I. Sutskever, and G. E. Hinton. Imagenet classification with deep convolutional neural networks. In *Advances in neural information processing systems*, pages 1097–1105, 2012.

[31] Z. Ma, Y. Yang, N. Sebe, K. Zheng, and A. G. Hauptmann. Multimedia event detection using a classifier-specific intermediate representation. *IEEE Transactions on Multimedia*, 15(7):1628–1637, 2013.

[32] M. Mazloom, E. Gavves, and C. G. Snoek. Conceptlets: Selective semantics for classifying video events. *IEEE Transactions on Multimedia*, 16(8):2214–2228, 2014.

[33] M. Mazloom, A. Habibian, D. Liu, C. G. Snoek, and S.-F. Chang. Encoding concept prototypes for video event detection and summarization. In *Proceedings of the 5th ACM on International Conference on Multimedia Retrieval*, pages 123–130. ACM, 2015.

[34] M. Mazloom, X. Li, and C. G. Snoek. Tagbook: A semantic video representation without supervision for event detection. *Multimedia, IEEE Transactions on*, 2016.

[35] M. Merler, B. Huang, L. Xie, G. Hua, and A. Natsev. Semantic model vectors for complex video event recognition. *Multimedia, IEEE Transactions on*, 14(1):88–101, 2012.

[36] T. Mikolov, I. Sutskever, K. Chen, G. Corrado, and J. Dean. Distributed representations of words and phrases and their compositionality. pages 3111–3119, 2013.

[37] G. A. Miller. Wordnet: a lexical database for english. *Communications of the ACM*, 38(11):39–41, 1995.

[38] D. Nister and H. Stewenius. Scalable recognition with a vocabulary tree. In *Computer vision and pattern recognition, 2006 IEEE computer society conference on*, volume 2, pages 2161–2168. IEEE, 2006.

[39] F. Perronnin and C. Dance. Fisher kernels on visual vocabularies for image categorization. In *Computer Vision and Pattern Recognition, 2007. CVPR'07. IEEE Conference on*, pages 1–8. IEEE, 2007.

[40] F. Perronnin, J. Sánchez, and T. Mensink. Improving the Fisher Kernel for Large-Scale Image Classification. In ., editor, *ECCV*, volume 6314 of ., pages 143–156, ., 2010.

[41] J. Philbin, O. Chum, M. Isard, J. Sivic, and A. Zisserman. Object retrieval with large vocabularies and fast spatial matching. In *Computer Vision and Pattern Recognition, 2007. CVPR'07. IEEE Conference on*, pages 1–8. IEEE, 2007.

[42] D. Qin, S. Gammeter, L. Bossard, T. Quack, and L. Van Gool. Hello neighbor: Accurate object retrieval with k-reciprocal nearest neighbors. In *Computer Vision and Pattern Recognition (CVPR), 2011 IEEE Conference on*, pages 777–784. IEEE, 2011.

[43] J. Sánchez and F. Perronnin. High-dimensional signature compression for large-scale image classification. In *Computer Vision and Pattern Recognition (CVPR), 2011 IEEE Conference on*, pages 1665–1672. IEEE, 2011.

[44] X. Shen, Z. Lin, J. Brandt, S. Avidan, and Y. Wu. Object retrieval and localization with spatially-constrained similarity measure and k-nn re-ranking. In *Computer Vision and Pattern Recognition (CVPR), 2012 IEEE Conference on*, pages 3013–3020. IEEE, 2012.

[45] B. Singh, X. Han, Z. Wu, V. Morariu, and L. Davis. Selecting relevant web trained concepts for automated event retrieval. In *IEEE International Conference on Computer Vision (ICCV), 2015*.

[46] J. Sivic and A. Zisserman. Video google: A text retrieval approach to object matching in videos. In *Computer Vision, 2003. Proceedings. Ninth IEEE International Conference on*, pages 1470–1477. IEEE, 2003.

[47] C. Sun, S. Shetty, R. Sukthankar, and R. Nevatia. Temporal localization of fine-grained actions in videos by domain transfer from web images. *ACM MM*, 2015.

[48] K. Tang, V. Ramanathan, L. Fei-Fei, and D. Koller. Shifting weights: Adapting object detectors from image to video. In *Advances in Neural Information Processing Systems*, 2012.

[49] A. Vedaldi and B. Fulkerson. Vlfeat: An open and portable library of computer vision algorithms. In *Proceedings of the international conference on Multimedia*, pages 1469–1472. ACM, 2010.

[50] S. Wu, S. Bondugula, F. Luisier, X. Zhuang, and P. Natarajan. Zero-shot event detection using multi-modal fusion of weakly supervised concepts. In *IEEE Conference on Computer Vision and Pattern Recognition (CVPR), 2014*.

[51] Z. Wu, Y. Fu, Y.-G. Jiang, and L. Sigal. Harnessing object and scene semantics for large-scale video understanding. In *Proceedings of the IEEE Conference on Computer Vision and Pattern Recognition*, pages 3112–3121, 2016.

[52] Y. Xia, X. Cao, F. Wen, G. Hua, and J. Sun. Learning discriminative reconstructions for unsupervised outlier removal. In *Proceedings of the IEEE International Conference on Computer Vision*, pages 1511–1519, 2015.

[53] Z. Xu, Y. Yang, and A. G. Hauptmann. A discriminative cnn video representation for event detection. *IEEE Conference on Computer Vision and Pattern Recognition (CVPR)*, 2015.

[54] G. Ye, Y. Li, H. Xu, D. Liu, and S.-F. Chang. Eventnet: A large scale structured concept library for complex event detection in video. In *Proceedings of the 23rd Annual ACM Conference on Multimedia Conference*, pages 471–480. ACM, 2015.

[55] Q. Yu, J. Liu, H. Cheng, A. Divakaran, and H. Sawhney. Multimedia event recounting with concept based representation. In *Proceedings of the 20th ACM international conference on Multimedia*, 2012.

[56] S.-I. Yu, L. Jiang, Z. Xu, Y. Yang, and A. G. Hauptmann. Content-based video search over 1 million videos with 1 core in 1 second. In *Proceedings of the 5th ACM on International Conference on Multimedia Retrieval*, pages 419–426. ACM, 2015.

[57] R. Yuster and U. Zwick. Fast sparse matrix multiplication. *ACM Transactions on Algorithms (TALG)*, 1(1):2–13, 2005.

[58] X. Zhang, Y. Yang, Y. Zhang, H. Luan, J. Li, H. Zhang, and T.-S. Chua. Enhancing video event recognition using automatically constructed semantic-visual knowledge base. *IEEE Transactions on Multimedia*, 17(9):1562–1575, 2015.

[59] L. Zheng, S. Wang, L. Tian, F. He, Z. Liu, and Q. Tian. Query-adaptive late fusion for image search and person re-identification. In *Proceedings of the IEEE Conference on Computer Vision and Pattern Recognition*, pages 1741–1750, 2015.

[60] L. Zheng, Y. Yang, and Q. Tian. Sift meets cnn: A decade survey of instance retrieval. *arXiv preprint arXiv:1608.01807*, 2016.

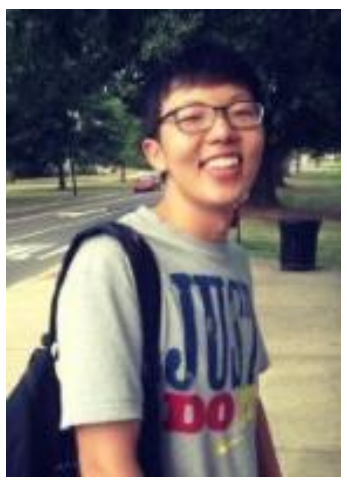

**Xintong Han** received the B.S. degree in electrical engineering from Shanghai Jiao Tong University, Shanghai, China, in 2013. He is currently pursuing the Ph.D. at the University of Maryland, College Park. His current research interests include computer vision, machine learning, and multimedia.

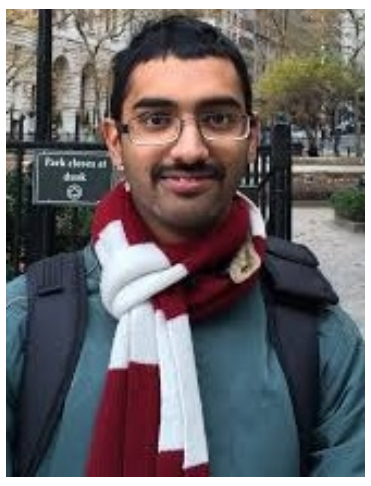

**Bharat Singh** Bharat Singh is a PhD student in Computer Science department at the University of Maryland, College Park. He received his Bachelor's and Master's degree in Computer Science from IIT Madras. His research interests are in Computer Vision and Machine Learning.

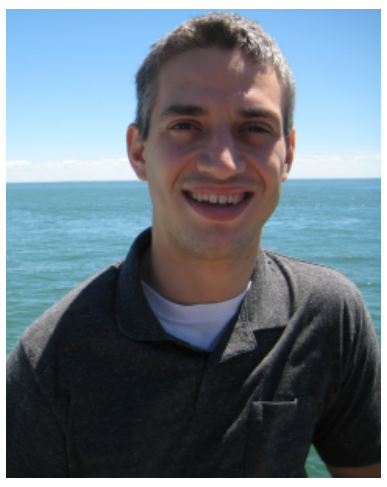

**Vlad I. Morariu** received the BS and MS degrees in Computer Science from the Pennsylvania State University in 2005 and the Ph.D. degree in Computer Science from the University of Maryland in 2010. He is currently an Assistant Research Scientist at the University of Maryland, associated with the Computer Vision Lab and the Institute for Advanced Computer Studies. His research interests are in computer vision, artificial intelligence, and machine learning, with a focus on representing, learning, and using visual knowledge to reason about images and videos in the presence of observation and model uncertainty.

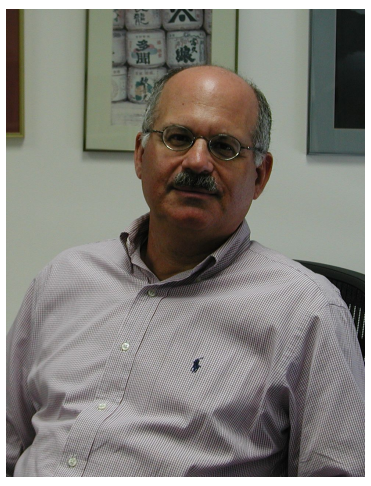

**Larry S. Davis** received the BA degree from Colgate University in 1970 and the MS and PhD degrees in computer science from the University of Maryland in 1974 and 1976, respectively. From 1977 to 1981, he was an assistant professor in the Department of Computer Science at the University of Texas, Austin. He returned to the University of Maryland as an associate professor in 1981. From 1985 to 1994, he was the director of the University of Maryland Institute for Advanced Computer Studies. He is currently a professor in the institute and in the Computer Science Department.